\documentclass[letterpaper, 10 pt, conference]{ieeeconf}  

\IEEEoverridecommandlockouts                              

\usepackage{graphicx}        

\usepackage{url}

\title{\LARGE \bf
SO-MRS: a multi-robot system architecture based on the SOA paradigm and ontology }

\author{Kamil Skarzynski$^{1}$,  Marcin Stepniak$^{1}$, Waldemar Bartyna$^{1}$, and Stanislaw Ambroszkiewicz$^{2}$,
\thanks{$^{1}$Waldemar Bartyna ({\tt\small wbartyna@gmail.com}), Kamil Skarzynski ({\tt\small kamil.skar@gmail.com}), and Marcin Stepniak ({\tt\small martinus.st@gmail.com}) -- Institute of Computer Science, University of Natural Sciences and Humanities, Siedlce, Poland}%
\thanks{$^{2}$Stanislaw Ambroszkiewicz -- Institute of Computer Science, Polish Academy of Sciences,  Jana Kazimierza 5, 01-248 Warsaw, Poland,
        {\tt\small sambrosz@ipipan.waw.pl}}
}

\begin{document}

\maketitle
\thispagestyle{empty}
\pagestyle{empty}

\begin{abstract}

A generic architecture for a  class of distributed  robotic systems is presented. The architecture supports openness and heterogeneity, i.e. heterogeneous components may be joined and removed from the systems without affecting its basic functionality. The architecture is based on the paradigm of Service Oriented Architecture (SOA), and a generic representation (ontology) of the environment. A  device (e.g. robot) is seen as a collection of its capabilities exposed as services. Generic protocols for publishing, discovering, arranging services are proposed for creating  composite services that can   accomplish complex tasks in an automatic way. Also generic  protocols for execution of composite services are proposed along with simple protocols for monitoring the executions, and for recovery from failures. A software platform built on a multi-robot system (according to the proposed architecture) is a multi-agent system. 

\end{abstract}

\section{Introduction}
\label{sec:intro}

The general assumption is that multi-robot system (MRS for short) consists of an environment, and of devices that operate in the environment and may change its state. A device may be considered as an element of the environment, and the device state may be subject of change, e.g. its position.   
The crucial  questions, for a MRS to be designed and developed, are: What is the purpose of the system? What kind of problems is it supposed to solve, or what class of tasks are to be accomplished in the system? 
If the system is dedicated to a fixed class of tasks, then the tasks as well as the methods for the task accomplishing may be hard-codded during  the design process.  


In the paper, a special kind of MRS is considered.  It is supposed that the  devices may be heterogeneous, and can be added to the system as well as be removed without affecting its basic functionality, i.e. the generic ability for task accomplishing. Hence, the class of the tasks is not fixed and depends on the joint capabilities of the  devices currently available in the system. Since such tasks can not be hard-coded in the system, there must be a language for the task specification. Intuitively, a task is an intention to change local state of the environment. That is, task consists of precondition and effect. Sometimes the precondition is not necessary.  
Precondition specifies initial local state of the environment, whereas the effect specifies the desired environment state after the task performance.  
So that, a formal representation of the environment (ontology) is needed. 
It is also supposed that the devices are not isolated, i.e.  there is  a minimum communication in the system in the form of (wireless) network. That is, each device has a network address and can receive and send  messages. 

Each device is autonomous and may provide some services (via its Service Manager) for a client (i.e.   human user or software application).  If a client has a task to be accomplished, it sends a request to the device. Then, the service may  accomplish the task, if it has enough resources and capabilities. 
Hence, each device provides some services that correspond to some types of elementary tasks the device may accomplish. The formal specification (expressed in a language of the common ontology, e.g. OWL-S \cite{OWL-S}) of the type of a service consists of a precondition and an effect. The service type must be published by a device (to be joined to MRS) to a Registry. Client may discover the service, and invoke it. This constitutes the essence of the Service Oriented Architecture (SOA) paradigm \cite{SOA} in Information Technology. 

%
Repository (the next component of MRS) is a realization of the common knowledge of the environment representation (ontology), and a storage of the current maps of the environment, i.e. instances of the ontology. Since the environment may be changed by devices, the maps must be updated.   

If a client wants to realize a complex task (a sequence or partial order of the elementary tasks), then some services, that may jointly accomplish the complex task, should be composed into a workflow (composite service).  An additional component of MRS is needed for doing so. It is called Task Manager, and it is responsible for constructing an abstract plan in the form of partial order of service types. Then, appropriate services should be arranged. Finally, the workflow is executed and its performance is monitored.  If a failure occurs (due to a broken communication or inability of a service to fulfill the arranged commitment), then failure recovery mechanisms must be applied.  
Simple mechanisms (in the form of protocols) consist in re-planning, and  changing some parts of the workflow in order to continue the task execution. 

To summarize, the software infrastructure (actually, a multi-agent system) built on MRS (for complex task accomplishing) consists of services exposed by Service Managers on devices (robots), service Registry, Task Manager, and Repository. The interactions between them are based on generic protocols for publishing, discovering, composing elementary services,  arranging, execution, monitoring and recovery from failures. Note that the basis for the protocols is a formal representation of the environment (ontology) that allows to specify local states of the environment, tasks, service types,  intentions, commitments, and situations resulting from  failures. Roughly, this constitutes the proposed architecture called Service Oriented Multi-Robot System (SO-MRS for short).  

SO-MRS architecture follows the hybrid approach based on additional infrastructure where main components of the infrastructure may be multiplied, i.e. in one MRS, there may be several independent Task Managers, Registries, and Repositories. Note that the presented approach is at higher level of abstraction than  Robot Operating System (ROS) that is usually used to implement services on the devices.  
The main contribution consists of simple universal upper ontology, MRS architecture, and generic protocols. 

The presented work is a continuation of Ambroszkiewicz et al.  (2010)  \cite{ambroszkiewicz2010multirobot}. 

\section{Motivations and related work} 
 
Rapid development and ubiquitous use of intelligent devices (equipped with sensors, micro-controllers, and connected to a network) pose new possibilities and challenges in Robotics and Information Technology. One of them is creating large open distributed systems consisting of heterogeneous devices that can inter-operate in order to accomplish complex tasks.  \emph{Ambient} \emph{Intelligence} (AmI), and \emph{Ubiquitous robotics} are currently extensively explored research areas. It is supposed that in the near future humans will live in a world where all devices are fully networked, so that any desired service can be provided at any place at any time. It is worth to notice the Intelligent Physical Systems research program by NSF, and Machine-to-Machine (M2M) standards promoted by International Telecommunication Union 
{\tt\small http://www.onem2m.org/}. 
AmI and Ubiquitous robotics require new information technologies for developing distributed systems that allow defining tasks in a declarative way by human users, and an automatic task accomplishing by the system.  Openness and heterogeneity of the systems are essential because of the scalability, see Di Ciccio et al. (2011) \cite{di2011homes}, and Helal et al. (2005) \cite{helal2005gator}. 

A lot of work has been done starting with the seminal papers by Fukuda et al. 1987 \cite{fukuda1987dynamically}, and by Asama et al.  1989 ACTRESS  \cite{asama1989design}. Several architectures were proposed for multi-robot cooperation:  
a pure swarm robotics approach using large numbers of
homogeneous robots, e.g. Matari{\'c} (1995) \cite{mataric1995issues},
and  Cao et al. (1997) \cite{cao1997cooperative}, a behavior-based approach without explicit coordination, e.g. ALLIANCE Parker (1998) \cite{parker1998alliance}, and a hybrid approaches, e.g. Distributed Robot Architecture (DIRA) Simmons et al. (2001) \cite{simmons2001first}. DIRA is closely related to SO-MRS, however, it was not fully developed. 
For a comprehensive overview, see Parker (2008) \cite{parker2008multiple}. 

Actually, the proposed SO-MRS architecture follows the idea of ASyMTRe-D and IQ-ASyMTRe \cite{tang2005distributed} \cite{parker2006building} \cite{zhang2013iq}. However, instead of sharing (by devices) mutually data from their sensors, SO-MRS is equipped with explicit common ontology as the basis for constructing generic protocols.

There are some other approaches that apply SOA paradigm, Semantic Web and Web Services technologies to multi-robot systems, like~the Ubiquitous Robotic Service Framework (URSF) project (2005) \cite{Ha} and (2007) \cite{Young}, Aiello et al. (2008) \cite{aiello2008our}, and  Kaldeli et al. (2013) \cite{kaldeli2013coordinating}. 
The project Service Oriented Device Architecture SODA Alliance \cite{SODA}, and its extension in the form of OASIS standard Devices Profile for Web Services (DPWS) \cite{DPWS}, is also of interest.  Device functionality is described there in the very similar way as it is done for Web services. 
For an extensive overview of SOA based robotic systems (from software engineering point of view) see de Oliveira (2015) \cite{deOliveira}.
Although these approaches are also based on SOA,  what makes the difference (in comparison to SO-MRS) is the lack of ontology, i.e. a \emph{common}  representation of the environment, and the language describing the representation. This very ontology is the necessary basis for constructing the generic protocols for automatic complex task execution and for recovery from failures. 
  
It is worth to notice that the following view presented by Parker (2003) \cite{parker2003current} is still up to date: {\em  ``A general research question in this vein is whether specialized architectures for each type of robot team and/or application domain are needed, or whether a more general architecture can be developed that can easily be tailored to fit a wider range of multirobot systems.''} 

It seems that there are still a lot of problems to be solved in the domain of multi-robot systems. 
Recent research directions are focused rather on software level.  
The player/stage project, Gerkey et al. (2003) \cite{gerkey2003player} developed  software level approach whereas the more abstract software independent level is needed. An abstract (however still unsatisfactory) approach was proposed by Kramer and Magee (2007) \cite{kramer2007self} as the software engineering point of view of the problem.  
The idea of Jung  and  Zelinsky (2000) \cite{jung2000grounded}, and Hugues (2000) \cite{hugues2000collective} of common grounded symbolic communication between heterogeneous cooperating robots is very close to the concept of common ontology, however,  it was not fully developed and not continued. 
An interesting approach to composite services (heterogeneous robot teams)  as temporal organizations with elements of recovery from failures was presented in Zhong  and DeLoach (2011) \cite{zhong2011runtime}. 

Let us also cite the view on the research on MRS by  Chitic, Ponage and Simonin (2014) \cite{chitic2014middlewares}: 
{\em 
``Despite many years of work in robotics, there is still a lack of established software architecture and middleware, in particular for large scale multi-robots systems. Many research teams are still writing specific hardware orientated software that is very tied to a robot. This vision makes sharing modules or extending existing code difficult. A robotic middleware should be designed to abstract the low-level hardware architecture, facilitate communication and integration of new software.''}

\section{Environment representation}
\label{sec:env_rep}
Classic representations of robotic environments  (see Thrun et al. (2002)  \cite{thrun2002robotic} for a comprehensive overview) are based on metric and topological approaches dedicated mostly to tasks related to navigation. Another approach, \emph{Spatial Semantic Hierarchy} (SSH), Kuipers (2000) \cite{kuipers2000spatial}, is based on the concept of cognitive map and hierarchical representation of spatial environment structure. There is some object-based approaches (see~ Vasudevan et al. (2007) \cite{vasudevan2007cognitive}) where the environment is represented as a map of places connected by passages.  Places are probabilistic graphs encoding objects and relations between them.  Anguelov et al. (2002) \cite{anguelov2002learning} proposed the environment representation  composed of two object hierarchies; the first one (called spatial) related to sensor data in the form of object images or occupancy grid, and the second one (called conceptual) related to some abstract notions of the representation. The recognition of places and objects consists in matching sensor data against the abstract notions. 
Recent work on semantic mapping is mainly focusing on perception and recognition techniques, see  
Pronobis et al. (2010) \cite{pronobis2010multi}, 
Pronobis and Jensfelt (2012) \cite{pronobis2012large}, 
and an extensive survey Kostavelis (2015) \cite{kostavelis2015semantic}. Also the project KnowRob, a knowledge processing system at RoboEarth 
\url{http://wiki.ros.org/roboearth}, explores this idea, see Tenorth and Beetz (2013)  \cite{tenorth13knowrob}. 
It seems that more abstract generic and simpler representation, in the form of upper ontology (common for humans and devices), is needed for MRS. 

In the Computer Science related to Robotics, the term ``ontology'' is equivalent to the ``general structure of the representation of a multirobot system environment''. The most popular definition of ontology was given by Tom Gruber (1993) ~\cite{gruber1993translation} in the following way: \emph{ontology is a specification of a conceptualization}.  Conceptualization is understood here as an abstract and simplified model (representation) of the real environment. It is a formal description of concepts (objects) and relations between them. Since the model is supposed to serve the interoperability, it must be common and formally specified, i.e. the definitions of objects and relations must be unambiguous in order to be processed automatically. 

\begin{figure}[htb]
\includegraphics[width=\linewidth]{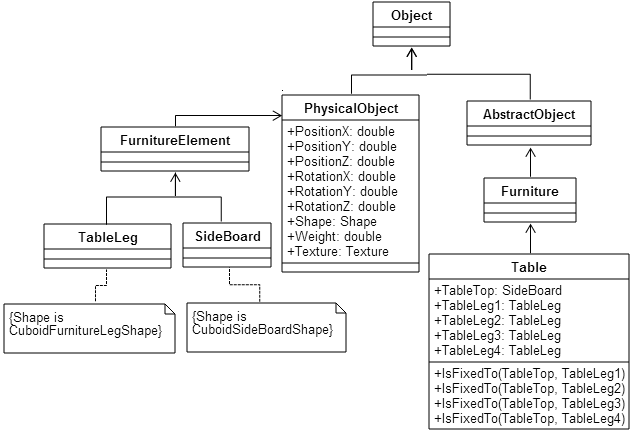}
\caption{Main object types}
\label{fig:objTypes}\vspace{-0.2cm}
\end{figure}

Two recent standards developed by groups of the IEEE RAS and addressing robot ontologies and map representation ({\small \url{https://standards.ieee.org/findstds/standard/1872-2015.html}} {\small\url{https://standards.ieee.org/findstds/standard/1873-2015.html}}) are closely related to upper ontology presented below. 
However, these ontologies are complex and include  specifications even for defining processes for task execution. Our ontology is extremely simple, and is at the higher level of abstraction, i.e. it abstracts from recognition of physical objects, and is focusing only on generic attributes of the objects that can be measured, recognized or evaluated.  
Although the proposed ontology is simple, it is generic (abstracting from implementation details), and sufficient together with simple and universal protocols (presented roughly below) to accomplish complex tasks in open and distributed heterogeneous multi-robot systems.  
In its general form it is an upper ontology, and consists of the following concepts:
\begin{itemize}
\item attributes that define  properties of object (e.g.: color, weight, volume, position, rotation, shape, texture, etc.,
\item relations that express dependencies between objects,
\item types of objects that specify object attributes, constraints on  attribute values, and relations between sub-objects,
\item object that is an instance of a type with concrete attribute values, sub-objects and relations between them.
\end{itemize}
In order to add a new type to the ontology one has to specify:
\begin{itemize}
\item parent type, i.e. the type that the  new type  inherits from,
\item list of attributes of the new type, 
\item list of types of obligatory sub-objects, i.e. types of objects that are integral parts of the type being defined, e.g. legs in the case of the type of table,
\item list of constraints specifying attribute values as ranges and/or enumerations, and  obligatory relations between sub-objects.
\end{itemize}
The type inheritance provides hierarchical structure supporting management of existing types as well as creation of new types. In the presented ontology the most generic type called \emph{Object} is inherited by two types: \emph{PhysicalObject} and \emph{AbstractObject} as shown in Fig~\ref{fig:objTypes}. The types are for separating physical objects that are directly recognizable by robots from abstract objects that are hierarchically composed from physical objects, relations between them, and  attributes. 

Descendants of \emph{PhysicalObject} type, that are leafs in the inheritance hierarchy tree, are called \emph{elementary} \emph{types}. They are described only by \emph{attributes} (\emph{simple} and/or \emph{complex attributes}) that can be recognized by robots. The type \emph{CuboidFurnitureLeg} is an example of an \emph{elementary} \emph{type}. It is defined by the following attributes: \emph{PositionX}, \emph{PositionY}, \emph{PositionZ}, \emph{RotationX}, \emph{RotationY}, \emph{RotationZ}, \emph{Shape}, \emph{Weight}, \emph{Texture}, and the  constraint: \emph{Shape is CuboidFurnitureLegShape}. \emph{Shape} is a \emph{complex attribute} consisting of its own attributes and their constraints. The constrains are important for object recognition, e.g.  attribute constrains of the \emph{FurnitureLeg} type are different than the constrains of the type corresponding to building pillars. 

The \emph{AbstractObject} branch consists of complex abstract types. Each such type is defined as a collection of types (complex and/or elementary), and relations between objects of these types. The type \emph{CuboidRoom} is an example of an abstract type. Internal structure of an object of this abstract type is composed of elementary objects such as walls, floor, ceiling, windows, and doors, as well as the relations between these objects. 

General structure of the proposed representation of the environment is defined as a hierarchy of types. Elementary type is defined as a collection of attributes with restricted ranges, whereas an abstract type is defined by some of the previously defined types (abstract and/or elementary), and relations between objects of these types. The type \emph{Building} consists of several other abstract types like storey, passages, rooms, stairs, lifts, etc.. 

The attributes and relations are the basic elements for creating representation, i.e. construction of object types. A particular object (as an instance of its type) is defined by specifying concrete values of its attributes, specifying its sub-objects (if it is of abstract type) and relations between them. Instance of the general structure (called also a map of the environment) is defined as a specification of an object of an abstract type, for example, of the type \emph{Building}.  In order to support an automatic map creating  and updating (by mobile robots), the  attributes must be recognizable and measurable by robot sensors. 

\section{Services} There are three kinds of services: 
\begin{itemize}
\item Physical services that may change situations in the physical environment.
 \item Cognitive services that can recognize situations described by formulas of the language of the ontology. 
\item Software services that process data.
\end{itemize}
A service interface consists of the following elements:
\begin{itemize}
\item Name of the type of service, i.e. name of an action that the service performs. 
\item Specification of the inputs and outputs of the service. 
\item The condition required for service invocation (precondition), and the effect of service invocation. 
\item Service attributes as information about the static features of a service, e.g. operation range, cost, and average realization time.
\end{itemize}

Precondition and effect are defined as formulas of  a formal language (OWL \cite{OWL}, or Entish~\cite{ambroszkiewicz2004entish}) describing local situations in the environment.  
Entish is a simplified version (without quantifiers) of the first order logic.  It has logical operators (\emph{and}, \emph{or}), names of relations (e.g., \emph{isIn}, \emph{isAdjacentTo}), names of functions (e.g., \emph{action}, \emph{range}), and variables.  A precondition formula is a description of the initial situation, and the effect formula is a description of the final situation.

\section{SO-MRS architecture}

\begin{figure}[thpb]\hspace{-0.2cm}
      \centering
      \includegraphics[width=\linewidth]{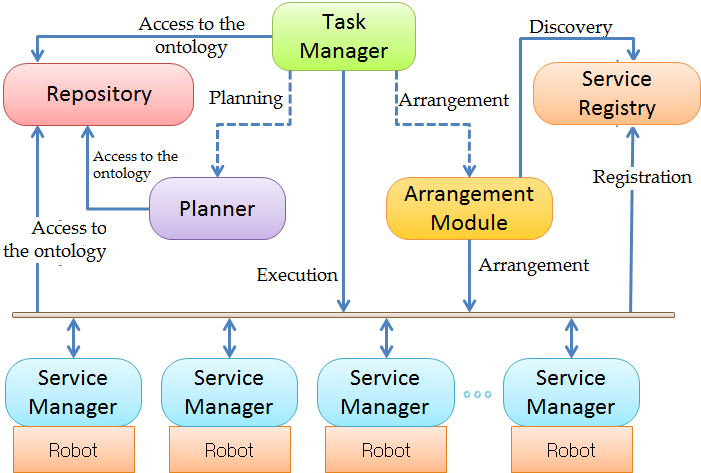}
      \caption{SO-MRS architecture}
      \label{architecture}\vspace{-0.2cm}
   \end{figure}

Figure~\ref{architecture} shows the proposed architecture of multi-robot system designed according to the SOA paradigm \cite{krafzig2005enterprise}.
The system components communicate with each other by using generic protocols. 
Repository stores ontology, and provides access to object maps for the  other system  components. It  has also a graphical user interface (GUI) for developing the ontology, and for its management.

Task Manager (TM for short) represents a client, and provides a GUI for  the client to define tasks, and to monitor their realization. The Planner provides abstract plans for TM, that are used  to construct a concrete plan (workflow) on the basis of information on available services (provided by Service Registry). The workflow is constructed  by arranging concrete services. 
Arrangement is performed by TM (via the Arrangement Module) by sending requests to services (in the form of intentions), and collecting answers as quotes (commitments).  TM controls the plan realization by communicating with the arranged services.  

Service Registry stores information about services currently available in the system. Each service, in order to be available,  must be registered  to Service Registry  via Service Manager (SM) that  is a  robot (device) interface for providing its services for an external client. In this case, TM acts as a client. SM controls the execution of subtasks delegated by TM, and reports the success or failures to TM. 

Task is defined (on the basis of the ontology stored in Repository) as a  logical formula that describes the initial situation (optionally) and the required final situation in the environment.  
For a given task, Planner returns abstract plans that, when arranged and executed, may realize the final situation specified by the task in question. An abstract plan is represented as a directed acyclic graph where nodes are service types and edges correspond to causal relationship between the output of one service and the input of another service. The relationship determines the order of arrangement, and then also the order of execution of a concrete plan (workflow) called also a business process.   
A concrete plan may also include handlers responsible for compensations, and failure handling to be explained below. 

\section{Protocol for failure handling and recovery}
 Since some ideas and methods are adopted from electronic business transactions, realization of a task is called {\em a transaction}.
 All services are invoked within a transaction that contains a dynamic set of participants.  
The transaction is successfully completed, if  the delegated task is accomplished.
Special transaction mechanism designed for handling failures has the following properties. 
\begin{enumerate}
  \item Failed services may be replaced by other services during task realization. 
  \item The general plan may be changed.
  \item The transaction ends either after successful completion of the task, or  inability to complete the task, or cancellation of the task.
\end{enumerate}

The classic meaning of the term {\em transaction} in Information Technology goes back to the  ACID properties of modifying a database. 
Long-running transactions avoid locks on non-local resources, use compensations to handle failures, potentially aggregate smaller ACID transactions (also referred to as atomic transactions), and use a coordinator to complete or abort the transaction. In contrast to rollback in ACID transactions, a compensation restores the original state or an equivalent one, and it is domain-specific, e.g.  for a failure when transporting a cargo by one robot, a compensation may be done by arranging another robot that can continue the transport to the destination, and charging (as a penalty) the owner of the first robot for the delay. 

   \begin{figure}[thpb]
      \centering
      \includegraphics[width=\linewidth]{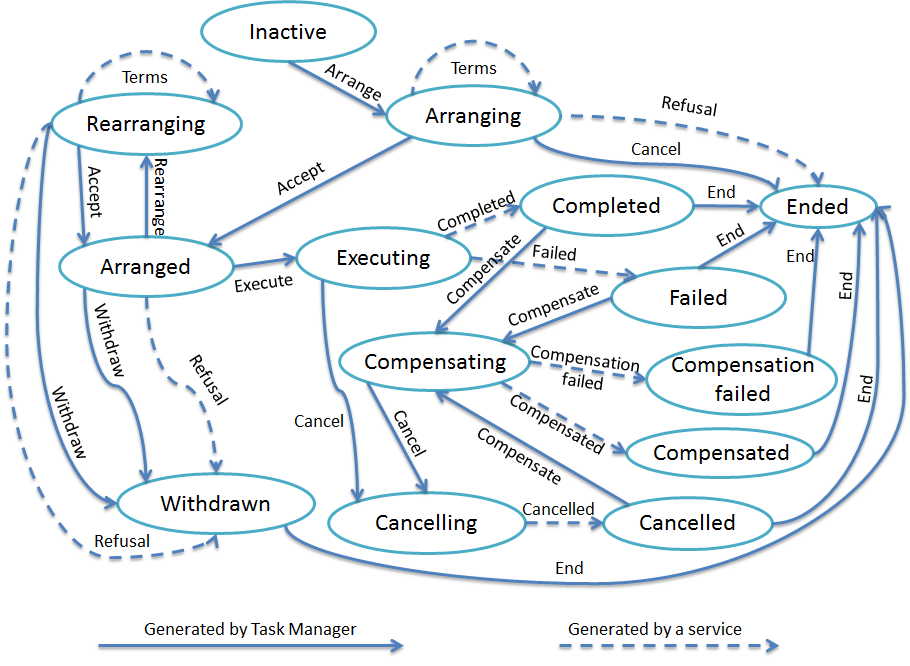}
      \caption{Transaction protocol state transition diagram}
      \label{protocol}\vspace{-0.3cm}
   \end{figure}

In distributed systems, a communication protocol specifies the format of messages exchanged between two or more communicating parties, message  order, and the actions taken when a message is sent or received. Based on the OASIS Web Services Transaction (WS-TX) standards  \cite{WS-TX}, a transaction protocol, called  Failure Recovery Protocol (FRP, for short), is proposed for multi-robot systems. FRP defines states of services, and types of messages exchange between Task Manager and services, see Fig.~\ref{protocol}. The messages are sent as SOAP 1.2 envelopes~\cite{soap2003}, and WS-Addressing~\cite{ws_addressing2004} specification is used. The message format consists of the header and the body. The header includes information about the sender, the recipient, the message type, the message and session identifiers, and the version of the protocol. The body contains data specific to the message type, e.g.  input data, output data, precondition, effect, or failure description.  All necessary data required for task execution and control are transmitted in the messages of  the transaction protocol.  
FRP allows TM to initialize particular phases of service invocation, monitor their progress, and perform additional actions, e.g. compensation. 
TM initializes the service execution by sending the required input data to the SM of the service. A service is invoked in accordance with the agreement made in the arrangement phase, and sends messages (via its SM and according to the protocol) to notify TM about the status of the performance of the delegated task.
 After successful execution, SM sends to TM the confirmation of  subtask completion, e.g. changing situation in the environment to the required one. TM can also stop the service execution before its completion. This may be caused by the task cancellation by the client, a failure during execution of other services in the plan (that cannot be replaced), or by changes in the environment making the current plan infeasible. 
Robot may not be able to successfully complete a task. In this case, its SM notifies TM by sending a detailed description of the problem. On this basis, TM can take  appropriate actions. If SM is not able to send such information, TM must invoke appropriate cognitive service (a patrolling robot, if available) to recognize the situation  resulting from the failure.  
Compensation is performed either after a cancellation of a subtask execution by a service, or after the occurrence of a failure that interrupts the execution. It is designed to restore the original state of the environment  before  the execution. Since restoring that situation is sometimes impossible, the compensation may change the situation resulting from the failure to a situation from which the task realization can be continued.    Note that even for simple transportation tasks (that seem to be simple) a universal failure recovery mechanism and corresponding compensations are not easy to design and implement. 
A concrete plan  should contain  predefined procedures for  failure handling and compensations. 
For an interesting approach and an extensive recent overview on the failure handling and recovery in robotic systems, see Hanheide et al. (2015) \cite{hanheide2015robot}. 

\section{Experiments verifying SO-MRS }
\label{sec:experiments}
SO-MRS architecture was implemented twice. The first implementation was done within the framework of Robo-enT project (2005-2008) with mobile robots Pioneer 3 (P3-DX). 

The Autero system (the RobREx project (2012-2015)) is the second implementation of revised and extended SO-MRS architecture with new version of the ontology and new protocols; see \url{http://www.robrex.ipipan.eu/about.php?lang=en} for the experiments. 
The system has been tested in a universal simulation environment implemented in Unity 3D. The class of tasks that can be accomplished in a real environment is always limited by the number and capabilities of available devices (robots). From the point of view of the proposed information technology (the architecture, ontology and protocols), the fact that the  test environment is simulated is irrelevant.
The simulation environment is generated {\em automatically (!)} from the contents of Repository. Service Managers are implemented as independent components that communicate with robots in the simulation environment via TCP/IP protocol.
 

\subsubsection*{Scenario 1 –- moving a jar from a cupboard to a platform}
The task was realized by a single \textit{TransferObject} service on a mobile robot with a gripper. The task is defined as:
\begin{itemize}
	\item precondition: {\em Jar002 isOn ?Shelf}
	\item effect: {\em Jar002 isOn Platform001}
\end{itemize}
In the arrangement phase, precondition and effect of the task are sent in an \textit{Arrange} message to the Service Managers that can provide the \textit{TransferObject} service. Two services of this type are registered in the system, so that, two Service Managers (representing these services) receive the same query. 
Service Managers respond with  \textit{Terms} messages, each of them  contains a commitment. 
 Additional service attributes (maximum service execution time, and price) are also specified. Service 2 has a shorter execution time whereas service 1 requests  much lower payment. So that, service 1 is selected for the task.
The SM of service 1 is notified by an \textit{Accept} message, whereas the SM of service 2 receives a \textit{Cancel} message.
The precondition is sent within an \textit{Execute} message initializing execution phase. After receiving this message, the Service Manager starts the task execution by moving the robot closer to the cabinet so that the object to be transported is within the range of the robot gripper. Then, the object is grabbed, the gripper is set to the transport position, and the robot approaches the platform on which the jar is put down.
Then, SM sends a \textit{Completed} message containing the description of the resulting situation.
In this scenario,  one service is needed for the task, so that, after the successful service execution, the task is considered as completed and transaction can be ended. The Task Manager does this by sending an  \textit{End} message to the Service Manager.
\subsubsection*{Scenario 1b –- failure}
\begin{figure}[tb]
\centering
\includegraphics[width=\linewidth]{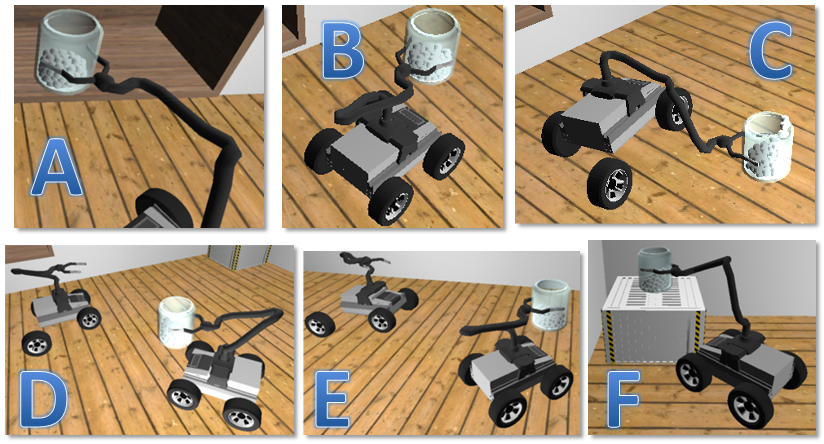}
\caption{"Moving a jar" scenario}
\label{sim_scenario1b}\vspace{-0.5cm}
\end{figure}
Failures might occur during the transportation, e.g. the robot drive was out of order. In such a situation, the robot has a control unit that can communicate with other system components. It also has an active gripper. Hence, it puts the transported object to the ground, and sends (via its SM) a \textit{Failed} message to the Task Manager containing information about the location of the jar (i.e. the formula {\tt \small Jar002.PositionX = 12.5 AND Jar002.PositionY = 1.3 AND Jar002.PositionZ = 7}). This allows the Task Manager to take an action in order to complete the task. In this scenario there is another service of the same (\textit{TransferObject}) type available. So that, TM arranges this service by passing, in the precondition, the situation (the new position of the jar) received from the damaged robot. The second service is executed, i.e. the operative robot goes to the position, picks up the jar, moves it to the destined position, and puts the jar on the platform.
Screen shots of Figure~\ref{sim_scenario1b} show the following steps:
A -- robot 1 approaches and grips the jar. 
B -- robot 1 transports the jar. C -- a failure;  the drive of robot 1 is out of order. 
D -- robot 2 approaches the jar and grips it.
E -- robot 2 transports the jar.
F -- robot 2 puts the jar on the target platform. 
\begin{figure}[htb]
\includegraphics[width=\linewidth]{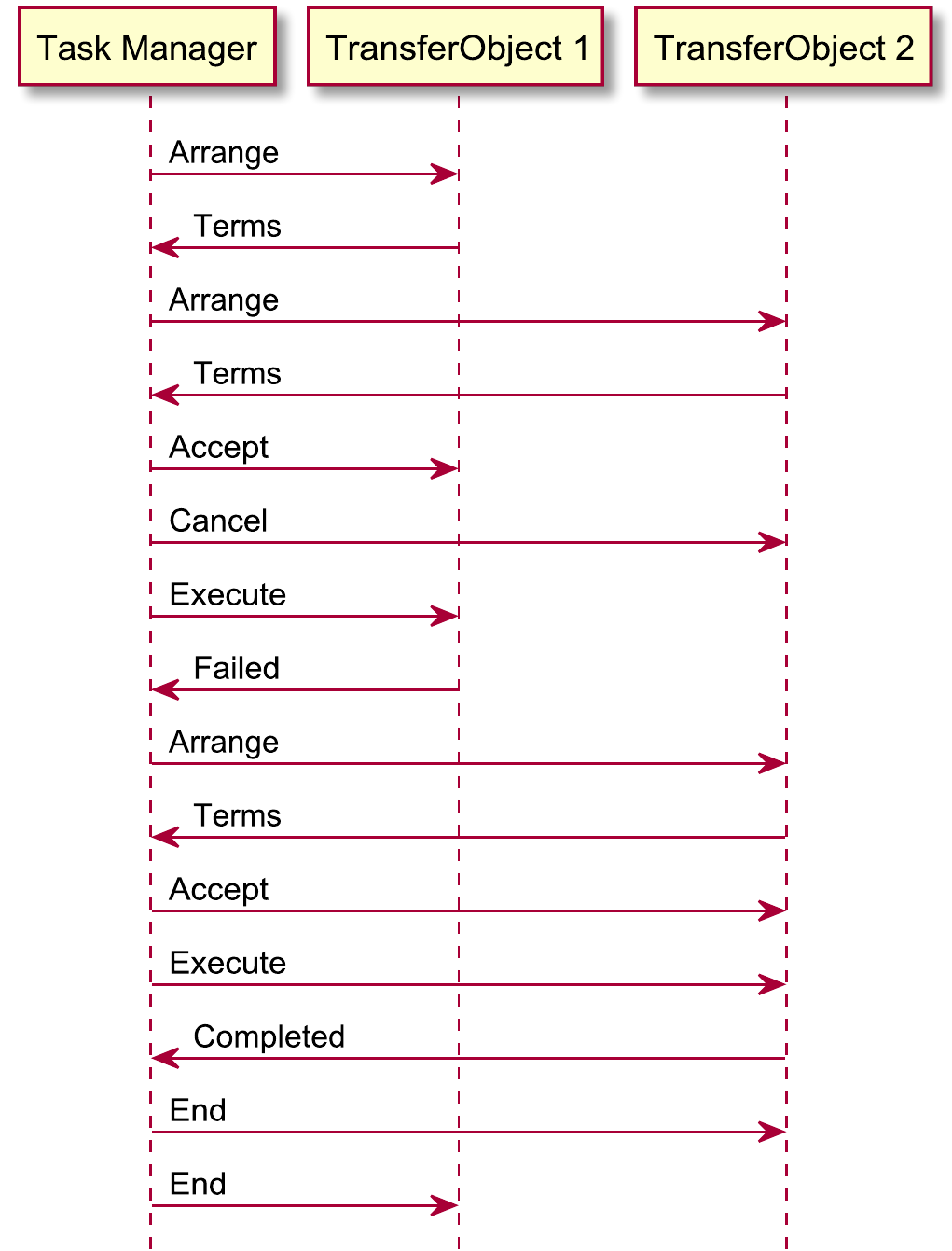}
\caption{Sequence of transaction messages in a transportation task with failure}
\label{Scenario1b}\vspace{-0.3cm}
\end{figure}
Figure \ref{Scenario1b} shows the complete FRP protocol message exchange sequence while performing the task in Scenario 1b. Failed \textit{TransferObject 1} service does not participate in the re-arrangement process. The second \textit{Arrange} message is sent only to \textit{TransferObject 2} service. The \textit{End} message is sent to both services, indicating the end of the transaction. More complex tasks (with failures during execution) were also tested.  
\\
{\bf Conclusion.}
The SO-MRS architecture is a proposal of a new information technology consisting of a generic environment representation (upper ontology),  specification of the system components, and generic protocols for realizing the system functionality, i.e. automatic accomplishing of complex tasks, along with a protocol for failure handling and recovery. 

\bibliographystyle{IEEEtran}
\bibliography{references}

\end{document}